\pdfoutput=1
\documentclass[11pt]{article}

\usepackage{acl}

\usepackage{times}
\usepackage{latexsym}

\usepackage[T1]{fontenc}
\usepackage[utf8]{inputenc}

\usepackage{microtype}

\usepackage{fixltx2e} % For subscript, \textsubscript
\usepackage{graphicx}
\usepackage{amstext} % Math text
\usepackage{caption}
\usepackage{subcaption} % Subfigures
\usepackage{xcolor} % Colors
\usepackage{lipsum}

\title{Letters From the Past: Modeling Historical Sound Change Through Diachronic Character Embeddings}
\author{Sidsel Boldsen$^{1}$ \and Patrizia Paggio$^{1,2}$\\
$^{1}$Department of Nordic Studies and Linguistics, University of Copenhagen \\ % , Copenhagen, Denmark
$^{2}$Institute of Linguistics and Language Technology, University of Malta \\ % , Msida, Malta
  \texttt{\{sbol,paggio\}@hum.ku.dk}\\}

\begin{document}
\maketitle
\begin{abstract}
%While the ability of word embeddings to model distributional semantic information has moved the field of study of Lexical Semantic Change (LSC), leading to formulation and testing of hypotheses governing this phenomena,
%PP 14-11 While the successful application of word embeddings to the computational modeling of Lexical Semantic Change has contributed to a growing interest for this field of studies, other aspects of language change -- concerning syntax, morphology and phonology -- have received less attention from the NLP community. 
While a great deal of work has been done on NLP approaches to %SB 14/3 changed to lowercase:
lexical semantic change detection, other aspects of language change have received less attention from the NLP community. 
In this paper, we %seek to study the process of %language
address the detection of sound change %and how it can be traced
through historical spelling. %By training diachronic character embeddings, we propose that a sound change, \textit{a}$\rightarrow$\textit{b / c}, can be captured by comparing the relative distance through time between its components, \textit{a} and \textit{b}. %While this hypothesis is confirmed
% PP, 14/3
%We propose that a sound change, \textit{a} $\rightarrow$ \textit{b} / \textit{c}, can be captured by comparing the relative distance through time between the distributions of the corresponding characters, \textit{a} and \textit{b}. We model these distributions using PPMI character embeddings.
We propose that a sound change can be captured by comparing the relative distance through time between the distributions of the characters involved before and after the change has taken place. We model these distributions using PPMI character embeddings.
We verify this hypothesis in synthetic data %, we also
and then test the method's ability to trace %a known
the well-known historical change of lenition of plosives %in Danish through
in Danish historical sources. %While we are able to
We show that the  models are able to identify several of the changes under consideration and to uncover meaningful contexts in which they appeared. %, and we also show how we can retrieve the contexts, \textit{c}, in which the sound change appears, providing promising results of corpus-based methods to study open questions as to the relative chronology of sound shifts and their geographical distribution of historical sound change.
The methodology has the potential to contribute to the study of open questions such as the relative chronology of sound shifts and their geographical distribution.
\end{abstract}

\section{Introduction}

%\textcolor{green}{TODO: Revisit + more references}

%\textcolor{green}{I think the introduction and motivation is okay, but I might make it a bit too complicated than it has to be. Also could it be made more appealing/exciting to the general CL audience?}

%Sound change has been studied since the birth of linguistics 
The study of sound change goes back to the beginnings of modern linguistics in early nineteenth century, when scholars such as Rask and Grimm started making hypotheses about the way sound changes over time, which in turn lead to the discovery of regular sound correspondences between ancient languages and the identification of cognates in modern ones \cite{murrayearly}.

Since spoken language from the past is not available, sound change in ancient languages must be deduced from written records by considering development in spelling through time.
In addition, while we may be able to see from the written records that a change did occur, less is known on the specific dynamics of the change. Details of these dynamics include knowledge of when the change started to appear, how long it took for it to be complete, what was the relative chronology of individual sounds in a larger shift, what was the geographical distribution of a change and so forth.

Due to the sparsity of linguistic evidence, detailed empirical studies of chronological sound change are difficult to conduct. This is especially the case for older stages of languages, where little written text was produced, and much of what did exist has %probably been lost. SB 14/11:
been lost in transmission.
However, as we move forward in history to the rise of bureaucracy, for example in medieval Europe, we see that an extensive amount of written records were made. Text from this period of time is interesting in the context of a study of sound change because it shows great variability in spelling patterns. %\textcolor{blue}{How many, how was spelling in this period, when did spelling start to stabilize.}\textcolor{red}{Not sure we need so much detail in the intro, maybe enough to say when spelling becomes more stable, but I don't know if it can be said in general, maybe better to speak about Denmark as a specific example.} 
With the digitalization of such archives\footnote{A list of available resources for different languages is provided in the Guide to Medieval Manuscript Research from the University of Chicago Library: \url{https://guides.lib.uchicago.edu/c.php?g=813534&p=5805534}.}, therefore,
new opportunities arise to apply computational methods to the study of sound change through written text. 

% Jeg synes ikke vi behøver tale om phylogenetics and linguistics. While the study of computational studies of sound change has mostly been studied using word lists, with phylogenetics as the main purpose, we wish to take a corpus-based approach, to try to answer some of the questions to how sound change starts to appear in writing. 
Considerable effort has already been devoted to the development of computational approaches aimed at discovering lexical semantic change (LSC) in historical corpora \cite{tahmasebi-etal-2018-survey}. However, change related to phonology, morphology and syntax has remained out of the spotlight in NLP research. In this study, we seek to bridge this gap as regards phonology. Inspired by the work on LSC, %which assumes that change in word meaning corresponds to changes of word embedding distributions, 
we propose a method whereby sound change is traced via spelling change in historical text and modeled by training diachronic character embeddings over text from different time periods.  %build upon a distributional hypothesis but for sounds by the training diachronic character embeddings to trace sound changes through historical spelling.

We start by reviewing previous approaches to the automatic detection of semantic shifts and spelling modification due to sound change. Then we formulate our hypothesis that a sound change %\texttt{A}  $\rightarrow$ \texttt{B} 
can be traced using diachronic distributional embeddings. While sound change is not completely analogous to word meaning change, we argue that similar methods can be used for both.
To verify our hypothesis, we conduct three studies on simulated sound change. First, we test the methods on the phonological environment of a simple artificial language. Then, we apply the same methods to a more complex scenario created by simulating sound change in a corpus of synchronic Danish text. Having established the suitability of the methods on these two datasets, we finally %PP 14/11 try to trace 
experiment with tracing a well-known sound change in real historical language data, again in Danish.

The implementation and datasets are available online\footnote{\url{https://github.com/syssel/letters-from-the-past}}.
%The goal is not only to detect sound change, but also to investigate questions such as relative chronology and change rate. SB 14/11:
%PP 14/11: I think it is better to say this in the conclusion. While the goal of this paper is to be able to detect sound change, we hope that such methods can proof corpus-based tool to investigate questions about the relative chronology, change rate and geographical distribution of sound change.

\section{Related Work}

%\textcolor{green}{TODO: Revisit the last part + more references}
The application of NLP methods to %PP 14/11 the automatic detection of lexical semantic change 
automatic LSC detection is already a rather well-developed subfield of NLP research \cite{tahmasebi-etal-2018-survey, kutuzov-etal-2018-diachronic}. In particular, the emergence of word embeddings as a viable way to model the  distributional hypothesis in semantics \cite{firth1957synopsis} has paved the way for an application of word embeddings to LSC modeling \cite{kim-etal-2014-temporal,hamilton-etal-2016-diachronic,eger-mehler-2016-linearity,Yaoetal2018}. Synchronically, the meaning of a word is characterized by word embeddings in terms of the contexts it appears in. LSC is captured by training word embeddings at different time points and comparing these distributions typically using cosine distance. 
%Embeddings trained on different partitions of a historical corpus must be aligned in vectorial space. Semantic change is usually computed based on the cosine distance between vectorial representations of the words under consideration as they are used in text from different temporal bins. 
%The comparison is often based on cosine distance.
% SB 14/3 added the following:
Recently, contextualized word embeddings have also been applied to the problem. While such models have the capability to capture change in distinct word usages, preliminary results suggest that traditional word embeddings are superior to the task of semantic change detection \cite{schlechtweg-etal-2020-semeval,montariol-etal-2021-scalable}.

One of the main issues in this comparison is the temporal alignment of dense embedding spaces. For example in the case of neural models, embeddings are initialized and trained stochastically, which means that separate runs -- on even the same data -- will yield different embedding spaces. Thus, work has focused on the development of methods to perform alignments to make dense embedding spaces comparable across time (see \citet{kutuzov-etal-2018-diachronic} for an overview).
As an alternative to neural embeddings, scholars have also used purely count-based measures, which are naturally aligned across dimensions. Normalization techniques are also applied, e.g. based on positive pointwise mutual information (PPMI) \cite{hamilton-etal-2016-diachronic, Yaoetal2018}.

Most studies of LSC do not rely on a control dataset against which to validate their conclusions. In \citet{dubossarsky-etal-2017-outta}, on the contrary, it is argued that any claims about putative laws of semantic change in diachronic corpora must be evaluated against a relevant control condition.  The authors propose a methodology in which a control condition is created artificially from the original diachronic text collection by reshuffling the data. %in several ways. ensuring that original and artificial corpora have the same vocabulary.
No systematic LSC is expected in the artificially developed control dataset. %The control corpus is valid if i. change scores diminish in the control condition compared to the original data, and ii. they are uniform across temporal bins in the control condition whereas more variation across bins is observed in the genuine data. Both these expectations are met, and the control corpus is subsequently used in the paper to investigate the effects of word frequency, polysemy and prototypicality on lexical semantic change. Such effects are modeled first in terms of correlations between change scores and the single factors (e.g. frequency), and then by means of a more complex linear mixed effects model. In both cases, the effects in question are studied by comparing the genuine corpus and the control shuffled one.
%The study argues based on the use of a control corpus that effects of word frequency, prototypicality and polysemy on LSC are not as large as previously claimed.

%Distributional phonology and mayer
%As for semantics (ref), 
The distributional hypothesis has also been proposed as an explanatory model within the domain of phonology suggesting that phonological classes are acquired through distributional information  \citep{chomsky1965sound, mielke2008emergence}. Driven by this hypothesis, recent work has focused on testing how distributional properties can be learned by phoneme embeddings (see \citealt{mayer-2020} for an overview). \citet{silfverberg-etal-2018-sound} investigated to what extent learned vector representations %(PPMI+SVD, Word2Vec, RNN) 
of phonemes align with their respective representations in a feature space in which dimensions are articulatory descriptors (e.g., $\pm$plosive). % (e.g., $\pm$consonant, $\pm$plosive, $\pm$dental). 
Recently, \citet{mayer-2020} has shown that phonological classes, such as long and short vowels, can be deduced from %PPMI+SVD embeddings of phonemes 
phoneme embeddings normalized using PPMI by iteratively performing PCA on candidate classes. %as a means of hierarchical clustering.

Thus, while the distributional hypothesis for phonology is well-established, one notable issue is the fact that the empirical evidence to study sound change is relatively inaccessible since it requires recorded speech or phonologically transcribed data.
%To account for the lack of empirical evidence available,
Simulation is therefore used as a tool for studying the underlying mechanisms of sound change by creating computational models based on linguistic theory \citep{wedelsimulation}. Through simulation, questions pertaining to e.g., what factors influence the (in)stability of vowel systems across generations \citep{deBoer2003Conditions} can be modeled by controlling the assumptions made by the model.
Work on simulation ranges from implementing theoretical approaches using mathematical models \citep{Pierrehumbert2001Exemplar, blythe2012s} to iterated learning and neural networks \citep{hare1995learning, begus2021deep}.

While the output of such models can be tested empirically on what we observe at a synchronic level, they are primarily theoretically driven.
In this paper, we wish to take a data-driven approach and utilize %the suggested methods for modeling lexical semantic change, where word embeddings are used to track sense change according to a distributional hypothesis. In order
some of the methods reviewed above to track historical sound change in writing. Rather than using word embeddings as done to model lexical change, we will use character embeddings, that are better suited to the task of sound change modeling. %Our methods also include the use of PPMI, simulation and control corpora. SB 14/11: Det beskriver vi i den næste sektion.

%\section{Distributional approach to the modeling of sound change}
\section{Modeling Sound Change}

%\textcolor{green}{TODO: Formuleringer er ikke så gode. Men jeg håber at det giver mening. Og notation i den sidste del skal opstrammes (+merges med hvad der er skrevet under experimental setup)}

%One main difference between modeling lexical semantic change and sound change is that in lexical semantic change, the form of the word remains the same, while the underlying distribution changes. Thus, change within LSC is usually modeled , in which change in

%PP, 12/11: I don't think we need to keep introducing.
%\textcolor{red}{Is it clear that LSC (and we) are talking about embeddings reflecting distributions? Should this somehow be repeated?}
%\textcolor{red}{ $\rightarrow$ Maybe just a short introduction about what we want to achieve in this small section. That is, we want to formulate an hypothesis how sound change can be traced distributionally through character embeddings. But we did say that in the end of the last section, so maybe it is clear.}
%PP, 12/11: I don't think it is a good idea to go into so much detail with LSC.
%This paper is not anout LSC but about sound change.

% SB 14/11: I am not sure I agree. I wrote this to introduce the different approaches to modeling change: pair-wise and individual displacement, which was introduced by Hamilton and that we we follow in our methodology.

Within the field of LSC detection, change in word semantics is traditionally measured by computing \textit{pairwise similarity} \citep{hamilton-etal-2016-diachronic} over a time series, ($t$, ..., $t+\delta$), in which a shift in the meaning of a word, $w_i$, can be measured by its relative distance to another word, $w_j$. In this way, hypotheses about specific shifts may be tested.
Another measure is \textit{semantic displacement}, in which semantic change for a given word is quantified by measuring its temporal displacement.
For both measures, looking at consecutive time steps provides a measure to the rate of change of a word -- in relation to another word, or independently.

While LSC is about meaning shifts of unchanged word forms, sound change is a change of form, i.e., a given phoneme changes to another one within certain contexts. We denote such a change \textit{a}
$\rightarrow$ \textit{b} / \textit{c}, where `c' stands for a given context.
%PP, 12/11: While such a change will be
%evident from looking at the displacement through time of
%either \textit{a} or \textit{b} , as both 
%their distributions are
%expected to change with respect to \textit{c}, such measure is
%nonspecific and would not only reflect the sound change but also other
%changing factors such as vocabulary.
While changes of either \textit{a} or \textit{b} will be reflected in
changes to their individual distributions
% SB 14/11 added
(\textit{displacement}), looking at them
independently of one another will not tell us whether one of the phonemes is becoming similar to the other.
%PP, 12/11: the vocablary problem exists also when we compare distributions, therefore I don't think it should be mentioned here.
%\textcolor{green}{Can we call the initial time t1? It is t1 in all the plots.}
%Instead,
Therefore, we suggest to look at the \textit{pairwise similarity}
between \textit{a} and \textit{b}. More specifically, given a time
series ($t_1$, ..., $t_n$), in which $t_1$ denotes a time before a
sound change was in effect and $t_n$ denotes a time where a sound
change is completed, we expect $b_{i}$ to \emph{move} towards
$a_{1}$ as $i \rightarrow n$, in other words to become similar to $a_{1}$,
%as
since it will begin to appear in the same
contexts.%, $c$, as $a_{0}$ as a sound change is completed.
%PP 12/11

As was noted earlier, sound is not accessible in historical text, to
which we would like to be able to apply our methodology.
% SB 14/3 added:
In historical text preceding spelling conventions, sound is assumed to be reflected in spelling. While detailed philological and linguistics analyses of written language can help to determine if a distinct spelling corresponds to a particular phoneme, or whether that spelling is rather a reflection of synchronic spelling variation \cite{minkova}, resources including such analyses are scarce. Thus, we chose
to use characters as a proxy for sound, and model sound change through
changes in the distance between pairs of character distributions.
In addition, before assuming that an observed decrease in the distance
between two such distributions reflects a real change, we
also want to see that the same decrease is not visible in a control
corpus in which no such change has indeed taken place.

\section{Experimental Setup}

In order to verify the hypothesis that sound change can be traced using distributional information with the methodology proposed above, we test whether we are able to trace simulated change in synthetic data. As a first synthetic setting, we restrict ourselves to track change in a synthetic language with simple phonotactics. In this way, we get a sense %PP 14/11 to 
of whether the proposed hypothesis works under perfect conditions,
i.e., one in which characters correspond with phonemes one-to-one. 
In the second synthetic setting, we seek to imitate the condition of tracing change in an orthographic setting by simulating change in a corpus of synchronic text in which character distributions interact with the noise added by spelling and lexicon.
In both synthetic settings, we compare the simulated change to a control setting where no change has occurred.  

%As a first step to verify the hypothesis that sound change can be traced using distributional information, we test if we are able to track a sound change in optimal conditions, i.e., in a limited phonotactic environment in which there is a one-to-one mapping between characters and phonemes. % Parupa

%As a second step, we want to track a well-known sound change in an orthographic setting in which phonotactic rules and possible phonological distributions interact with the noise added by spelling and lexicon.
%In order to do this, we collect a corpus of synchronic text which we divide into separate partitions where we simulate a sound change in different degrees such that each partition represents a different stage of this change. % UD-Danish

%The purpose of developing models of sound change for these two relatively limited datasets is to test the general viability of the methodology before it is applied to real diachronic language data.

Finally, we will test the hypothesis on real data. %PP 14/11 to see if we are able 
Our goal is to trace the lenition after vowels of voiceless plosives, \textit{p t k}, to their voiced counterparts, \textit{b d g}, in historical Danish. While this change is believed to %PP 14/11 be 
have initiated around the beginning of the 14\textsuperscript{th} century, details about the relative chronology of the series and geographical distribution of the change are difficult to account for \citep{frederiksen-2018-dansk-sproghistorie}.
%PP 14/11 By training character embeddings on historical sources from the periods following the beginning of the change, we are interested to see if we are able to trace this change in writing according to the proposed hypothesis.
Therefore, in an attempt to discover interesting patterns of this change, we train character embeddings on historical sources from the periods following the time when the change is believed to have started.
%PP 14/11 Following 
As we did for the synthetic data, and again following \citet{dubossarsky-etal-2017-outta}, we also introduce a control setting to test the significance of the observed changes.%PP 14/11 changes that are observed.

%We want to see if the embeddings are able to capture this change through changes of distribution of the corresponding characters, and maybe give us an indication of when the change started and from which of the three characters (e.g., if \texttt{/p/} shifted before the other consonants in the series, and thus initiated the change).

%In all three sets of experiments, we compare the results obtained on the target corpus with those from a control corpus in which the postulated sound change has not happened. In the following sections we describe each dataset (both target and control versions) and the way the character embeddings were built and compared across time periods.

%\begin{figure*}[h]
%  \begin{subfigure}[b]{0.5\textwidth}
%    \includegraphics[]{figures/parupa.pdf}
%    \centering
%    \caption{}
%  \end{subfigure}
%  \begin{subfigure}[b]{0.5\textwidth}
%    \includegraphics[]{figures/uddanish_wc.pdf}
%    \centering
%    \caption{}    
%  \end{subfigure}
%  \caption{}
%  \label{fig:pca_sim}
%\end{figure*}

\subsection{Data}
\paragraph{Parupa} is an artificial language introduced by \citet{mayer-2020}.
It is characterized by %PP 14/11 having 
a small phonological inventory\footnote{\textit{C}: /\textit{p t k b d g r}/ \textit{V}: /\textit{i e u o a}/}, and simple phonotactic rules for how sounds combine:
\begin{itemize}\itemsep0em
  \item only \textit{CV} syllables are allowed
  \item /\textit{p t k}/ occur before high vowels, /\textit{i u}/
  \item /\textit{b d g}/ occur before non-high vowels,\\
      /\textit{e o}/
  \item  only /\textit{b p}/ occur word-initially
  \item /\textit{r}/ occurs before all vowels
  \item all consonants can occur before /\textit{a}/
\end{itemize}

We created five corpora of Parupa consisting of 20,000 words each using the Hidden Markov Model provided by \citet{mayer-2020}\footnote{\url{https://github.com/connormayer/distributional_learning}}:
While the first corpus, \texttt{parupa\textsubscript{1}}, preserves the phonotactic rules listed above,
the remaining four include a sound change, \textit{p} $\to$ \textit{b} /\_ \textit{u, i}
% SB 14/3: Addded further description to explain what an underscore means in a context:
\footnote{The underscore indicates the position of the changing element, i.e., \textit{p} changes into \textit{b} when preceding \textit{u} or \textit{i}. This notation using an underscore to indicate position of the changing element will be used throughout the rest of the paper.}%In this and subsequent sound change examples, we will not use phoneme notation since our models manipulate characters and not phonemes directly.}}, SB 14/11: I think this is clear from the introduction to the section
, which %halfway to completed (realized in half of the cases) in \texttt{parupa\textsubscript{2}} and 
happens gradually (linearly) and is fully completed in \texttt{parupa\textsubscript{5}}.
Additionally, we created five control corpora (one for each of the target ones and with the same vocabulary) which do not include any simulated sound change.
%two separate corpora for the control study, \texttt{parupa\textsubscript{2} (control)} and \texttt{parupa\textsubscript{3} (control)}, which do not include the above sound change. \textcolor{red}{Add why two control corpora are needed.}
Each of the corpora consists of $50,000$ words.\\

\paragraph{The Danish UD treebank}
To collect a corpus of synchronic language, we extracted the training sentences from the Danish UD treebank \citep{johannsen2015universal}. From this collection of sentences, we extracted five sub-corpora (\texttt{UD-Danish\textsubscript{1-5}}) consisting of $\mathtt{\sim}$16,000 words each, in which we simulated a sound change, \textit{g $\to$ k / V\_\{V \# t\#\}}\footnote{i.e., \textit{g} between vowels, word-final after vowel, or after vowel preceding word-final \textit{t}. The latter condition was created in order to capture adverbial forms of adjectives ending in -\textit{g}.}. As done in the case of Parupa, the sound change was simulated gradually, with linear increase in change probabilities (i.e.,  $0, 0.25, 0.50, 0.75, 1$). To create the control condition, we also kept a version of the sub-corpora where no change was simulated. The five control versions are thus identical to the five target corpora in terms of vocabulary and distributions, except for the simulated change.\\

\paragraph{Historical spellings of geographical names}
\textit{Danmarks Stednavne} is a on-going lexicographic book series %PP 14/11 seeking to create
creating a register of  geographical names in Denmark. The register also serves as a philological resource by listing attestations of the names coming from various historical resources.
% SB 14/11 added:
For example, the entry for \textit{Copenhagen} includes over 700 historical attestations listed by date\footnote{e.g., \textit{Kopmanahafn} (1247), \textit{Køpmannehafn} (1249), \textit{Kiøpnehaffn} (1388), \textit{Kiøbendehaffn} (1429).}.
In addition to the printed volumes \citep{danmarksstednavne}, geographical names and their connected metadata (e.g., geographical location and historical attestations) have been digitized, and can be found in an online edition\footnote{\url{https://danmarksstednavne.navneforskning.ku.dk}} which comprizes over $210,000$ names and $900,000$ historical attestations. To study the lenition of /\textit{p t k}/, we extracted historical attestations of names ranging from the 12\textsuperscript{th} to the 18\textsuperscript{th} century. Using the attestation before the 14\textsuperscript{th} century as a reference to the time before the change was initiated ($t_1$), we divided the list of names into bins of half a century to track the development of character embeddings through time.
% SB 14/3 moved the following out of footnote as we have space:
The choice of bin size is an important methodological consideration when tracing language change \citep{kutuzov-etal-2018-diachronic}. From a philological perspective, 50 years correspond to two generations of writers (`spellers'), which is considered a realistic bin size to track development of spelling in writing.
This provides us with eleven sub-corpora with $31,000$ ($\pm15,000$) %number of samples 
name tokens on average.

In order to create a control setting, we %PP 14/11 created 
generated a corresponding number of sub-corpora by stratifying the names with respect to their date of attestation, %SB 14/3 added:
corresponding to the `shuffle' approach suggested by \citet{dubossarsky-etal-2017-outta}.
In this approach, no diachronic change is expected to be observed, as attestations are distributed evenly across bins based on their original date of occurrence.

\subsection{Character Embedding Model}
% Turn argumentation around. We want something simple, and therefore we use PPMI. Then they do not have to be aligned, which is known to cause noise. As a plus, they are easy to interpret, and works well with our hypothesis.

To represent characters in a distributional space, we use PPMI embeddings. %TODO: Definition.
Contrary to dense embeddings, these are easy to interpret and when compared across different initializations, they are naturally aligned, so we do not introduce noise caused by the alignment process.

Using the implementation by \citet{mayer-2020}, the sliding window is directional, and thus we distinguish contexts preceding and following the target character. While this directionality is neglected when creating PPMI word embeddings, the direction matters %PP 14/11 a lot when dealing with character representations. This has something to do with the intuition behind the distributional hypothesis (that contexts depend on position). \textcolor{red}{Explain a bit better}
when using character embeddings to test the intuition behind the distributional hypothesis, in which direction in a context is meaningful.

% SB 14/3 added the following:
The context window is chosen according to the conditioning of the change aimed to be captured: For Parupa, the simulated change is conditioned on only one character, and thus for this experiment we applied bigrams. For UDDanish, we applied trigams as the change is conditioned by two characters (the preceding and succeeding).
For the tracking of lenition in Danish, the condition of the change is expected to be similar to the one we simulated in the synthetic setting of UDDanish. However, to ensure we capture enough context, in this case we expand the model to using 4-grams.

\subsection{Measuring Change}
We measure sound change in terms of a decrease in the distance between
two character distributions over time. In other words, given two
character distributions A and B corresponding to any two
phonemes /\textit{a}/ and /\textit{b}/, we should see that $distance(
A^{(1)}, B^{(n)})$ gets smaller for greater values of {\it n} if there
is a change $A\rightarrow B$.

While most studies use cosine distance to measure the difference between distributions \cite{kutuzov-etal-2018-diachronic}, we chose to use
Euclidean distance as it directly reflects our hypothesis by taking the sum of differences in each dimension (context).

For each of the corpora being investigated, we use the R
software \cite{R_project} and the `effects’
package \cite{FoxWeisberg2019} to build linear regression models that
predict the distributional distance between two sounds per temporal
interval in the target and the control versions of the corpus. The
advantage of employing linear regression in this case is that we can
test the effect of multiple factors as well as their interaction. %PP 14/3 \footnote{See also \cite{shoemark-etal-2019} on the advantages of using linear regression in semantic change detection.}
In our case, the distance between the two sounds being investigated is the
dependent variable, and we want to predict the main effects of
temporal interval and corpus as well as the interaction between
them. To argue that there has been a sound change across time, there
must be a significant effect of temporal interval on distance. In
addition, we would like to see an interaction between this effect and
the effect of the corpus variable in that the change should be absent,
or at least significantly smaller, in the control corpus.

\section{Results}

% Description of the change that we observe in the two first studies

% How we evaluate the change w.r.t. pair-wise distance

%\begin{figure*}
%  \includegraphics[]{figures/ds_all_wc.pdf}
%  \caption{\textcolor{green}{Jeg ved ikke hvor meget information %graferne over kontrolstudiet viser. Måske skulle vi hellere have %billederne af pair-wise distance, som jeg sendte dig i mailen.}}
%\end{figure*}

\begin{figure}
     \centering
     \begin{subfigure}[b]{0.4\textwidth}
         \centering
         \includegraphics[width=\textwidth]{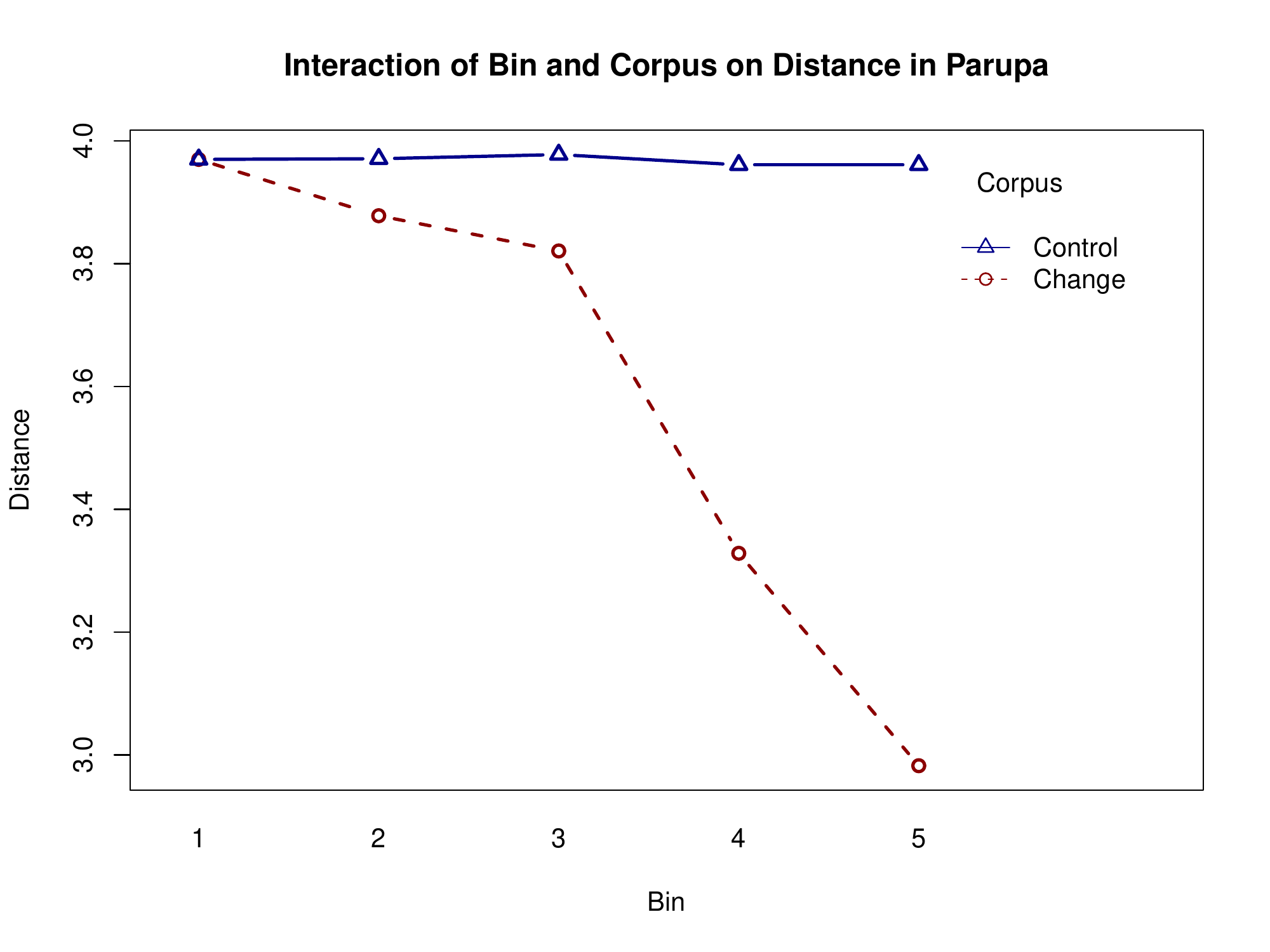}
         \caption{}
         \label{}
     \end{subfigure}
     \hfill
     \begin{subfigure}[b]{0.4\textwidth}
         \centering
         \includegraphics[width=\textwidth]{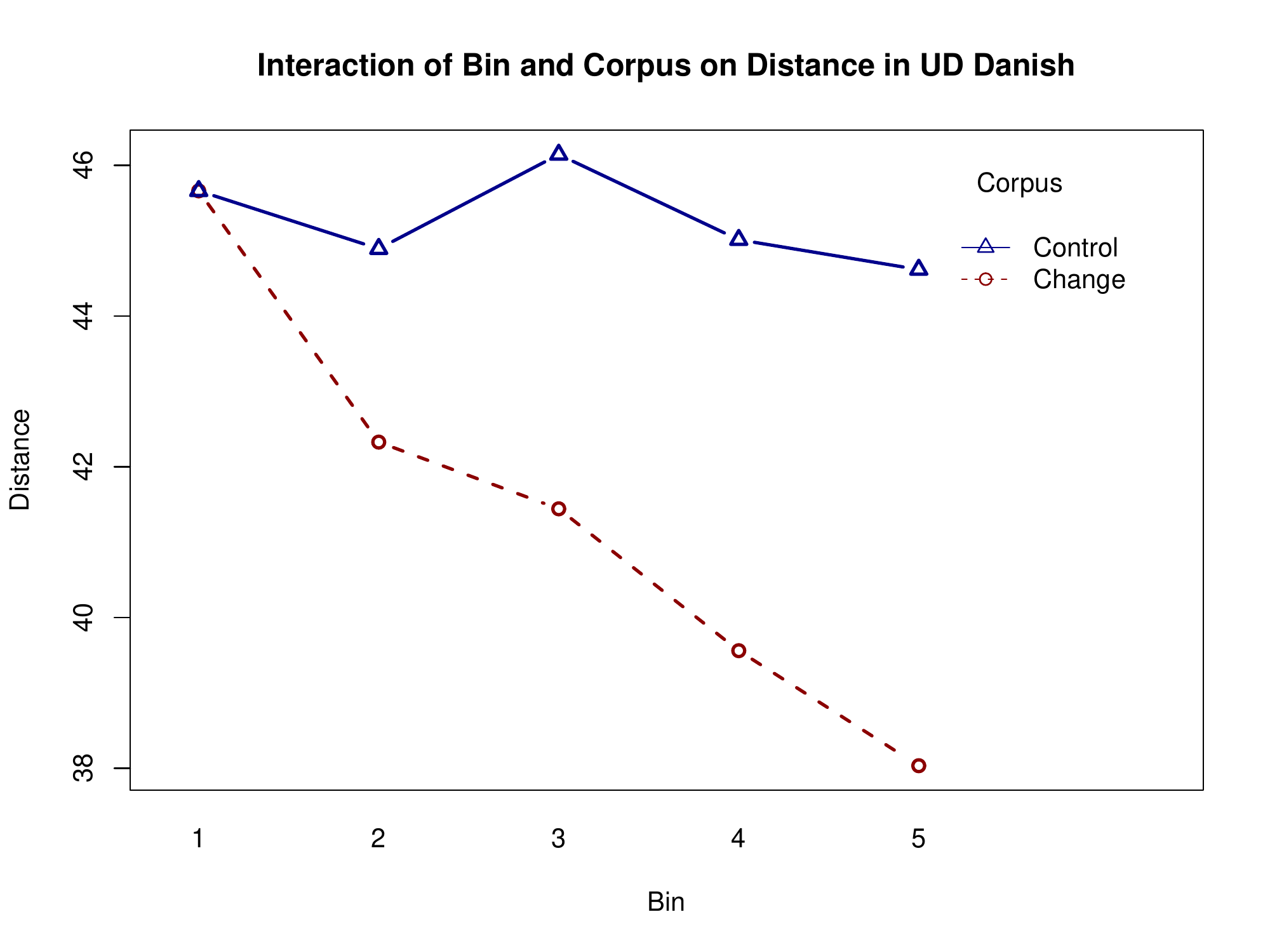}
         \caption{}
         \label{}
     \end{subfigure}
        \caption{Interaction of Bin and Corpus on Distance in Parupa (a) and the Danish UD treebank (b)}
        \label{fig:interactions}
\end{figure}

\begin{figure}
     \centering
     \begin{subfigure}[b]{0.4\textwidth}
         \centering
         \includegraphics[width=\textwidth]{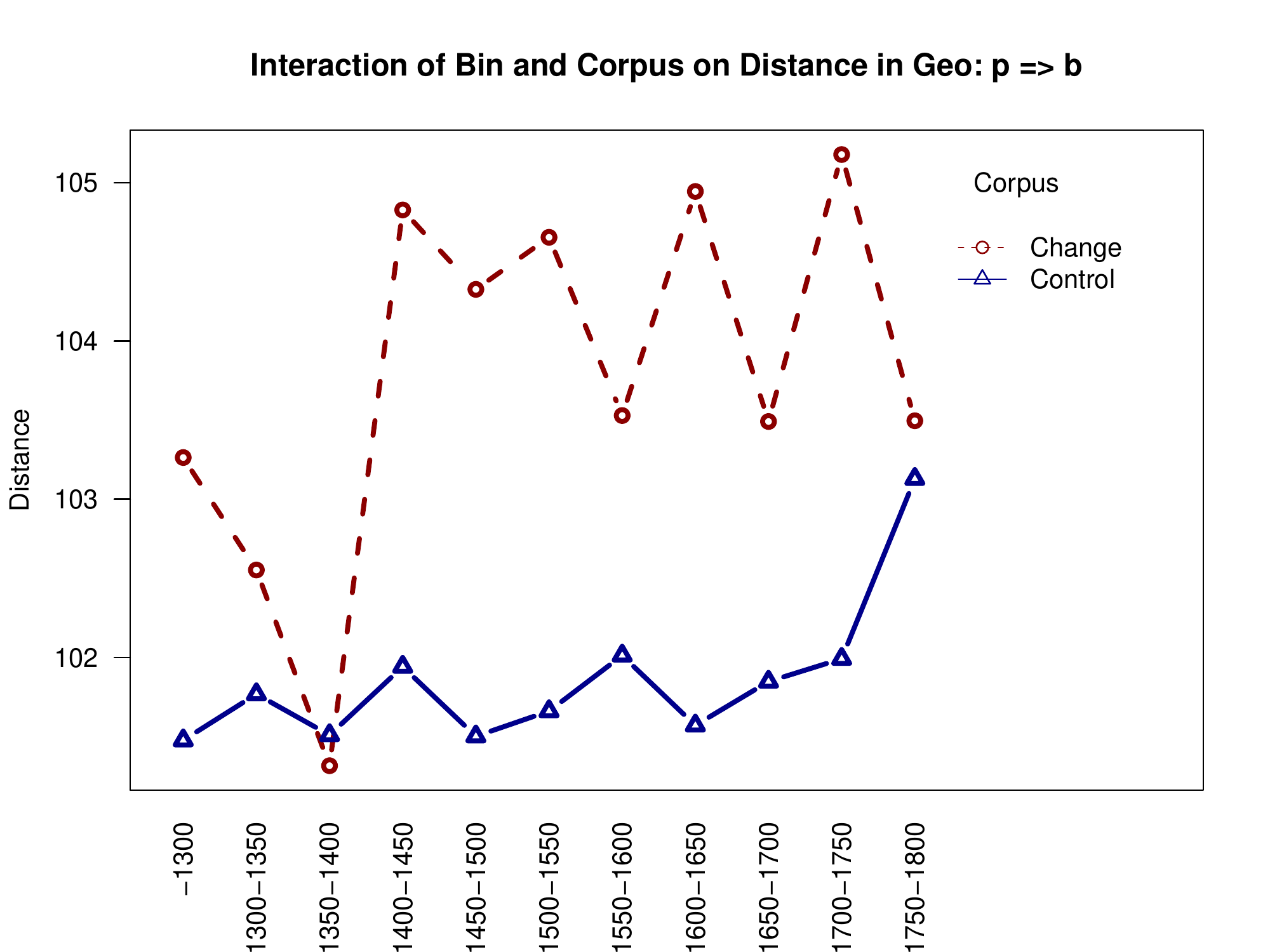}
         \caption{}
         \label{}
     \end{subfigure}
     \hfill
     \begin{subfigure}[b]{0.4\textwidth}
         \centering
         \includegraphics[width=\textwidth]{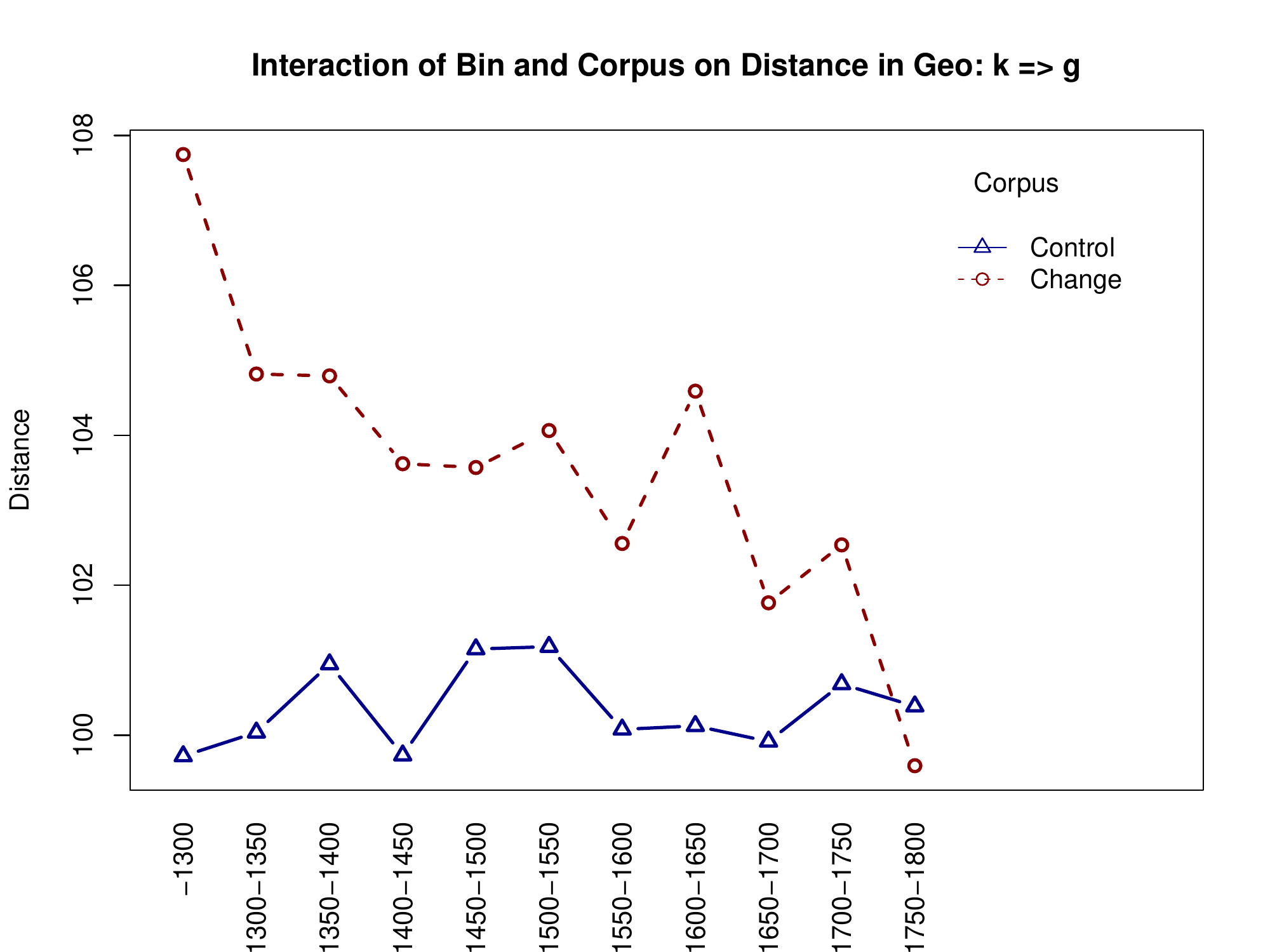}
         \caption{}
         \label{}
    \end{subfigure}
     \begin{subfigure}[b]{0.4\textwidth}
         \centering
         \includegraphics[width=\textwidth]{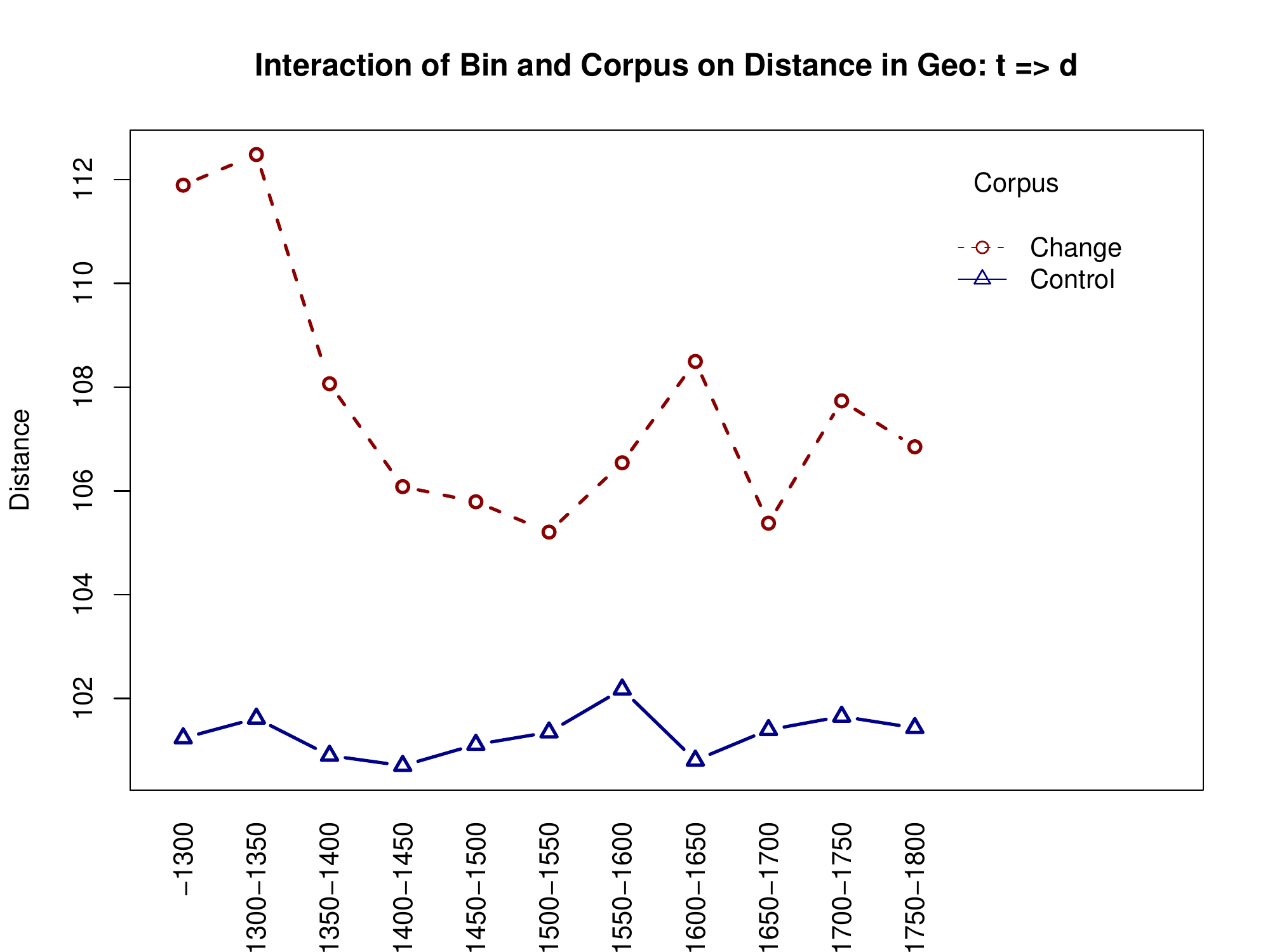}
         \caption{}
         \label{}
     \end{subfigure}
        \caption{Interaction of Bin and Corpus on Distance in the Danish Geographical Names: Looking at p $\rightarrow$ b (a), k $\rightarrow$ g (b) and t $\rightarrow$ d (c)}
        \label{fig:interactions2}
\end{figure}

\begin{table*}
\centering
\begin{tabular}{llccc}
\hline
&\textbf{Effect} & \textbf{Estimate} & \textbf{Std. Error} & \textbf{p-value} \\
\hline
&(Intercept) &4.35 & 0.12 &  $<$.001 ***\\
Parupa & Bin & -0.25 & 0.04 &  $<$.001 *** \\
&Control & -0.38 & 0.17  &  0.07 \\
&Bin:Control  & 0.25 & 0.05 & $<$.01 **\\
\hline
&(Intercept)   &     46.81  &   0.67 & $<$.001 ***\\
UD Danish & Bin  &  -1.80  &   0.20  & $<$.001 ***\\
& Control &    -0.96 &    0.94  & 0.35\\  
& Bin:Control &   1.60 &    0.28 &  $<$.01 **\\ 
\hline
&(Intercept)   & 102.82  &   0.53 & $<$.001 ***\\
Geo Names & Bin  &  0.16  &   0.08  & 0.054 \\
p $\rightarrow$ b & Control & -1.50 & 0.75 & 0.06\\  
& Bin:Control & -0.07 & 0.12 &  0.53\\ 
\hline
&(Intercept)   & 106.77  &   0.59 & $<$.001 ***\\
Geo Names & Bin  &  -0.53  &   0.09  & $<$.001 ***\\
k $\rightarrow$ g & Control & -6.55 & 0.84 & $<$.001 ***\\  
& Bin:Control & 0.55 & 0.12 &  $<$.001 ***\\ 
\hline
&(Intercept)   & 110.23  &   0.99 & $<$.001 ***\\
Geo Names & Bin  &  -0.42  &   0.15  & $<$.01 **\\
t $\rightarrow$ d & Control & -9.13 & 1.41 & $<$.001 ***\\  
& Bin:Control & 0.46 & 0.21 &  $<$.05 *\\ 
\hline
\end{tabular}
\caption{Coefficients of linear regression models predicting increase of distance between the investigated sounds in two simulated corpora.}
\label{tab:effects1}
\end{table*}

Table~\ref{tab:effects1} shows the results of the linear regression models we developed to test whether any evidence of sound change discovered in the target corpora, where sound change is either simulated or historically present, stands the comparison with the control corpora. The `intercept' estimate corresponds to the distance predicted between the two sounds being investigated in the initial temporal interval. The `Bin' estimate shows by how much the distance is expected to change for every temporal interval. A negative effect means that the distance between the two sounds is becoming smaller. The `Control' effect shows the predicted change to the initial Intercept in the control corpus %PP 14/3 added
(this corresponds to the effect of the corpus variable), and finally `Bin:Control' shows the interaction between temporal bin and corpus type. 

%PP 14/3 formula added
In the R implementation, each of the models is expressed in terms of the following equation:
\begin{equation}
model=lm(Distance \sim  Bin*Corpus)
\end{equation}
In both corpora where change is simulated, there is a significant effect of temporal interval. This is expected given the fact that gradual change has been induced in the data. %PP 14/3 The effect of the control corpus on the initial sound distance is significant for Parupa but not for UD Danish. More importantly
For both corpora, the effect of the control corpus on the initial sound distance is not significant. Importantly, the interaction between the effect of the temporal bin and the control corpus is significant in both cases. The interaction supports the hypothesis that we see a pattern of change in the simulated corpora that is significantly different compared to the control data. The interactions are shown in the plots in Figure~\ref{fig:interactions}. 

Turning to the results for the Danish Geographical Names corpus, while the models show significant effects of Bin, Control and interaction between the two for the \textit{k} $\rightarrow$ \textit{g} and the \textit{t} $\rightarrow$ \textit{d} changes, %PP 14/3 only the effect of Bin is significant 
no significant effects are found for the \textit{p} $\rightarrow$ \textit{b} change. When we look at the corresponding interaction plots in Figure~\ref{fig:interactions2}, we see that the distance between {\emph p} and {\emph b} in the corpus %PP 14/3 seems to increase rather than diminish (as also shown by the positive Bin effect), and to do so in a rather non-linear way.
decreases in the third bin to then increase and finally slightly decrease again in a non-linear way.
The changes displayed in the plots in (b) and (c), on the contrary, follow the expected trend: The observed consonant is moving towards its voiced version in the real corpus but not in the control.

\section{Discussion}

The results from the two simulation studies suggest that sound change can be traced with our proposed methodology of measuring %PP 14/11 the time-wise distance between its components. 
the distance between pairs of character distributions over time.
%PP 14/11 First, we show 
We showed this both in a simplified setting (Parupa), and  %PP 14/11 in a noisy orthography of
in the orthographically noisy environment provided by synchronic Danish data (UD Danish).

The main assumption in these simulation studies was that change %PP 14/1 is linear
could be modeled linearly. However, as discussed by scholars, change is %PP 14/3 most 
often not linear, %PP 14/3 but rather follows 
and can follow an s-shaped curve through a community \citep{denison_2003}. 
In a study of semantic lexical change based on synthetic data, \citet{shoemark-etal-2019} experiment with the injection of changes the probabilities of which vary linearly or logarithmically, and find that regression in general provides reasonable results in spite of being sensitive to outliers and of producing a certain amount of false positive results. It also performs better than a non-parametric measure like Kendall's $\tau$.
%PP 14/3 In a similar study on synthetic data, nevertheless, \citet{shoemark-etal-2019} %PP 14/1 show while modifying the change rate influence the recall of simulated change, 
%showed for LSC detection that tracing the change under a linear assumption, such as ours, still performs well. %PP 14/1 A bit vague, so I have rephrased it. While we have no reason as to why the measures on diachronic character embeddings should behave differently from those of word embeddings, it would be interesting to investigate in relation to sound change, if some types of change are more difficult to track than others.
The results obtained in our study seem to confirm %PP 14/7 this finding in the case of
the usefulness of linear models to detect sound change even though one of the cases of lenition targeted in the Danish Geographical Names corpus could not be modelled.

%PP 14/3 Moving on to on the results on the tracing of lenition in historical sources, 
Focusing on our results on the tracing of lenition, then, %PP 14/11 while we were able to identify a change from  $/t \;k/\rightarrow /d\;g/$, the results do not tell us much about what patterns the model picked up.
we were able to identify a change from  $/t \;k/\rightarrow /d\;g/$.
% SB 14/3: Added the following discussion points
However, an important thing to note in regards to the control setting for these results is how it diverges from the synthetic settings, which we initially used as a verification of the proposed hypothesis to trace sound change.
There, the variation in vocabulary was the same in the simulated and the control settings. In this case, however,
vocabulary variation is lower in our control setting due to the shuffling of the name attestations.
As a consequence, the control setting does not properly test the possible confounding effect of vocabulary within the proposed methodology.
Therefore, we proceeded to evaluate what types of contexts the model picked up.

\begin{table}[]
  \centering
    \begin{tabular}{ccc}
    \hline
    2-gram              &   Slope    & Pearson's {\emph r} \\
    \hline
        \texttt{\_u} &  -0.20  &  -0.88 \\
        \texttt{\_i} &  -0.19  &  -0.89 \\
        \texttt{i\_} &  -0.07  &  -0.94 \\
        \texttt{a\_} &  -0.06  &  -0.98 \\
        \texttt{o\_} &  -0.06  &  -0.89 \\
    \hline
    \end{tabular}
        \caption{Analysis of the simulated change from p to b in Parupa. Five most important dimensions after filtering bigrams with respect to Pearson's {\emph r} (<-0.2) and p-value(<0.05). The table is ordered by slope. `\#' indicates word boundaries.}
        \label{tab:patternsparupa}
\end{table}

 \begin{table}[]
 \centering
     \begin{tabular}{ccc}
    \hline     
     3-gram              &   Slope    & Pearson's {\emph r} \\
    \hline
        \texttt{li\_} &  -0.71  &  -0.93 \\
        \texttt{i\_e} &  -0.64  &  -0.89 \\
        \texttt{i\_t} &  -0.59  &  -0.93 \\
        \texttt{di\_} &  -0.58  &  -0.98 \\
        \texttt{a\_e} &  -0.57  &  -0.96 \\ 
    \hline
     \end{tabular}
     \caption{Analysis of the simulated change from g to k in synchronic Danish. Five most important dimensions after filtering trigrams with respect to Pearson's {\emph r} (<-0.2) and p-value(<0.05). The table is ordered by slope. `\#' indicates word boundaries.}
     \label{tab:patternsuddanish}
 \end{table}

\begin{table}[]
 \centering
     \begin{tabular}{ccc}
        \hline
        4-gram              &   Slope    & Pearson's {\emph r} \\
        \hline
        \texttt{rvi\_ }  &     -0.49  &   -0.85 \\
        \texttt{æ\_er }  &     -0.42  &   -0.78 \\        
        \texttt{sii\_ }  &     -0.40  &   -0.71 \\
        \texttt{m\#a\_ }  &     -0.40  &   -0.81 \\
        \texttt{oli\_ }  &     -0.39  &   -0.84 \\  
        \texttt{an\_h }  &     -0.32  &   -0.80 \\
        \texttt{ara\_}  &     -0.31  &   -0.62 \\
        \texttt{n\_ga}  &     -0.29  &   -0.82 \\
        \texttt{vi\_\# }  &     -0.29  &   -0.70 \\
        \texttt{is\_a}  &     -0.29  &   -0.73 \\
    \hline
     \end{tabular}
     \caption{Analysis of the change from k to g in historical records of geographical names. Ten most important dimensions after filtering 4-grams with respect to Pearson's {\emph r} (<-0.2) and p-value(<0.05). The table is ordered by slope. `\#' indicates word boundaries.}
     \label{tab:patternsstednavne}
 \end{table}
To get a sense of this, instead of looking at the euclidean distance for the full embedding, we %PP 14/11 reran 
ran linear regression %PP 14/11 added a few things
on the target data looking at differences 
between character distributions for each dimension. %PP 14/11 By considering the slope and Pearson's \textit{r} of the dimensions, we retrieve the patterns that elicited the most change in our previous measures
We then extracted the patterns corresponding to the dimensions showing significant differences and considered those with the highest Pearson's \textit{r} coefficient (Tables~\ref{tab:patternsparupa}-\ref{tab:patternsstednavne}).

Starting with the resulting patterns for Parupa and UD Danish, in both cases we are able to identify the exact contexts where the change was simulated: In Parupa before \textit{i}/\textit{u} and in the UD Danish corpus, between vowels and in the frequent suffix -\textit{ig(t)} (although the end-of-word is not captured due to n-gram size restrictions).
% SB 15/3 added
For Parupa, it is worth noting how the model captures patterns after vowel as well. This position is only implicitly involved in the conditioning of the simulated change, and the slope correspondingly less steep.

Moving on to the tracing of sound change in real data, we focus our analysis on \textit{k} $\rightarrow$ \textit{g}, which showed the greatest change. Considering the patterns, \texttt{rvi\_} and \texttt{vi\_\#}, these are connected to the the word \textit{vig} `inlet', commonly used as a suffix in the formation of geographical names in Danish. Descending from a Proto-Germanic word with final -\textit{k} (\textit{w\={\i}kwan} `to give way; to turn (away)', compare German \textit{weichen} `id.' and Dutch \textit{wijken} `id.' \citep{EtymologicalDictionaryofProtoGermanic}), the suffix is in early sources attested with a -\textit{k}: For example, out of the six written sources of the geographical name \textit{Rørvig} before the 14\textsuperscript{th} century (corresponding to bin 1-3 in our study), four were written with a -\textit{k}, while in later sources forms with -\textit{g} became predominant, with the latest attestation of -\textit{k} appearing in 1465.
Many of the patterns can be attributed to spellings related to similar changes\footnote{Danish \textit{sig} `bog; mire' from Old Danish \textit{sik}, compare Norwegian and Swedish (dialectal) \textit{sik} \citep{danmarksstednavne}}\textsuperscript{,}\footnote{
 Danish \textit{ager} `field' from Proto-Germanic \textit{akra}, compare English \textit{acre} and Swedish \textit{åker} \citep{EtymologicalDictionaryofProtoGermanic}.}. However, in the case of \texttt{n\_ga}, \texttt{is\_a} and \texttt{an\_h} these are not immediately interpretable. In the case of \texttt{oli\_}, this pattern is most likely related to the word \textit{bolig} `home;dwelling'. This word, however, does not have a comparable ancestor with -\textit{k}, and the change has to be explained as reflecting later innovation, namely beginning trend of using \textit{bolig} in name formations among younger attestations.

%PP 14/11 This stresses 
This latter example is related to an important issue in %PP 14/11 the wider field of modeling 
language evolution: When language changes through generations, we also observe shifts in culture. Different types of `data drift' are in fact discussed by \citet{hamilton-etal-2016-cultural} %PP 14/11 within 
in the context of LSC. The authors %PP 14/11 show how these 
suggest that they may be modeled independently of each other  by means of different measures of change. %PP 14/11 This 
The effect of cultural change has yet to be discussed for sound change. However, it is an important discussion, since phonology, when looking at it from a corpus-based perspective, is not only governed by phonotactic constraints, but also a by-product of %lexicon and 
word usage, which is in turn dependent on cultural patterns.

In this respect, another important point to note about the retrieved patterns -- both from the simulation of UD Danish and the tracing of \textit{k} $\rightarrow$ \textit{g} -- is that many of them reflect derivational or inflectional suffixes, and are thus characterized by %PP 14/11 occurring frequently 
high frequency of occurrence across word forms. 

%PP 14/3 rephrased and moved to the end.
%While %PP 14/11 such an 
%the observation that frequent patterns are more easily captured %PP 14/11 might 
%may seem trivial, %--  --  it is an important trait of the model, namely that it 
%it cannot be ignored that the model may be less sensitive to infrequent patterns.\footnote{
%Whether frequency could explain the lack of evidence for observing \textit{p} $\rightarrow$ \textit{b} is to be investigated further. Germanic \textit{p} descends from Proto-Indo-European (PIE) *\textit{b}, which has a special place in the PIE phoneme inventory, being the black sheep that some scholars do not believe to have existed due to its few attestations. Thus, the attestations of Germanic \textit{p} most often come from loan words and are %PP 14/11 often 
%not seen in morphemes. Thus the evidence for \textit{p} $\rightarrow$ \textit{b} is inherently scarcer.}

While the observation that frequent patterns are more easily captured may seem trivial, lack of sufficient evidence may nevertheless be the reason why we cannot model the  \textit{p} $\rightarrow$ \textit{b} change. Germanic \textit{p} descends from Proto-Indo-European (PIE) *\textit{b}, which, however, has a special place in the PIE phoneme inventory and is considered a sort of black sheep that some scholars do not believe to have existed due to its few attestations. In fact, the attestations of Germanic \textit{p} most often come from loan words and are not seen in morphemes. Thus the evidence for \textit{p} $\rightarrow$ \textit{b} is inherently scarcer than for the other two consonant pairs we have investigated. Further investigation of this sound change could be carried out by means of additional simulations, or more detailed analysis of the obtained character distribution, and is left for the future.

% SB 15/3: Ændret i forhold til reviewers kommentarer + nye resultater. Og flyttet til slut i afsnittet: Jeg synes at det afsnit som du flyttede ned til sidst egentlig hang meget godt sammen med det hvor det stod. Og på samme tid blev dette afsnit taget lidt ud af sammenhæng.

A final observation on the identified patterns is that the model fails to generalize across synchronic variation in spellings. For example, we see that a spelling with \textit{ii} is treated alongside spelling with a single \textit{i}. While this type of variation could to some extent be accounted for by treating it as an independent variable,
such a solution would have consequences for our experiment design in that we use PPMI weighting on raw n-gram counts. %PP 14/1, without any SVD or the like. 
This method enabled us to interpret the exact %PP 14/11 mechanism of the model, finding 
inner workings of the model and find the contexts in which a change has happened. If we had used neural models for example, in which characters are represented by dense embeddings, similar characters would %PP 14/11 share
have shared similar representations, %PP 14/11 which could imitate how, e.g., linguistcs make generalisation about sound changes happening after \textit{vowel}, such issues might be mitigated, but also would have been harder to uncover. 
thereby perhaps allowing the model to generalise e.g., to sound change occurring after a \textit{vowel}. %PP 15/11 However, interpreting the results would probably have been harder.
In this study, we wanted to privilege explainability, but dense representations should be explored in the future.
% PP 15/11 Moved part of the next paragraph to the conclusions.
%Utilizing neural methods would be a next step, and when doing so, it would be interesting to test how they compare to simple methods such as %PP 14/11 this
%the one used in this paper, %PP 14/11 to see how they tackle traits such as 
%particularly as regards infrequent patterns that could be %PP 14/11 generalised upon 
%captured across word forms.

%\section{Evaluation}
 
%\newpage

%\textcolor{red}{How are these measures related to the multiple linear regression models that you ran? Can we use the same terminology? E.g., for slope?}
%\textcolor{red}{I'm not sure whether the effect is calculated in exactly the same way as the slope in correlation analysis, so I would leave things as they are. May for the revision, I need more time to look into it.}

\section{Conclusion and Future Work}
% Whether this could be used to study the underlying dynamics of a sound change. E.g., dynamics in a shift of a series. How fast it was realized in writing. But we measure the effects linearly.

In this paper we presented a novel method %PP 15/11 of modeling 
for the modeling of sound change through the use of diachronic character embeddings. Sound change is modeled in terms of increasing similarity between character distributions across time intervals. %PP 15/11 While testing proposed method on synthetic data, we also show how they apply to tracing 
The proposed method was tested on synthetic data with promising results, and then applied to a real world scenario with the goal of tracing the lenition of /\textit{p t k}/ $\rightarrow$ /\textit{b d g}/ in Danish by looking at spelling in historical sources. 
%PP 15/11 When considering the contexts in which the change is detected, the method is able to detect well-known patterns, and thus, it proves as a promising method to a corpus-based approach for the studying the underlying mechanisms of sound change such as the relative chronology and geographical distribution of sound shifts.
The method was able to detect the changes for two of the sound pairs, and also to point at specific contexts of occurrence that influenced the changes.
% SB 14/3: Added following consideration about our results
However, our evaluation showed that the proposed models were sensitive to variation relating to vocabulary. To what extent such variation is responsible for the occurrence of false positives has yet to be investigated.

For scholars interested in sound change, there are a number of important open questions, such as the relative chronology and geographical distribution of sound shifts. Although we have not addressed these questions here, we believe our methodology can be further developed in ways that would allow to do so, e.g., by adding geographical location as an additional factor in the models. Both issues would constitute interesting avenues for future research.

In this paper we have used purely count-based methods. While this approach enables us to directly interpret the results of the models, it also suffers from %PP 15/11 the lack of generalising across contexts. 
its inability to generalise across contexts.
This drawback motivates experimenting with %PP 15/11 using neural methods to test to compare how well they are able to utilize 
neural methods that make use of dense character representations, to test whether they can make similar generalisations as done by historical linguists, particularly as regards infrequent patterns that could be captured across word forms.

\section*{Acknowledgements}
We would like to thank Bo Nissen Knudsen and David Caspersen Yousif for helping to prepare and making the data from Danmarks Stednavne available for the study.
Also, a great thanks to the anonymous reviewers for their helpful comments and suggestions.

The first author is supported by the project \textit{Script and Text in Time and Space}, a core group project funded by the Velux Foundations.

% Bibliography
%\bibliography{custom}
\bibliography{article}
\bibliographystyle{acl_natbib}

%\appendix

%\section{Example Appendix}
%\label{sec:appendix}

%This is an appendix.

\end{document}